%% file: main.tex
\documentclass[letterpaper, 10 pt, journal, twoside]{IEEEtran}
\IEEEoverridecommandlockouts
\usepackage{cite}
\usepackage{amsmath,amssymb,amsfonts, dsfont}
\usepackage{algorithm, listings}
\usepackage{algpseudocode}
\usepackage{graphicx}
\usepackage{textcomp}
\usepackage{xcolor}
\usepackage{mathtools}
\usepackage{blindtext, graphicx}
\usepackage{subcaption}
\usepackage{multirow}
\usepackage[normalem]{ulem}
\usepackage{diagbox}
\usepackage{hyperref}
\usepackage[absolute,overlay]{textpos}

\def\argmin{\mathop{\arg\min}\limits}	%
\def\argmax{\mathop{\arg\max}\limits}	%

\def\BibTeX{{\rm B\kern-.05em{\sc i\kern-.025em b}\kern-.08em
    T\kern-.1667em\lower.7ex\hbox{E}\kern-.125emX}}

\begin{document}
\markboth{IEEE Robotics and Automation Letters. Preprint Version. Accepted November, 2020} {Nardari \MakeLowercase{\textit{et al.}}: Urquhart Tessellations}


\onecolumn
\textcopyright 2020 IEEE. Personal use of this material is permitted. Permission from IEEE must be obtained for all other uses, in any current or future media, including reprinting/republishing this material for advertising or promotional purposes, creating new collective works, for resale or redistribution to servers or lists, or
reuse of any copyrighted component of this work in other works.
\vspace{5pt}

Pre-print of article that will appear in the IEEE Robotics and Automation Letters (RA-L).

\vspace{5pt}
Please cite this paper as:
\vspace{5pt}

G. V. Nardari, A. Cohen, S. W. Chen, X. Liu, V. Arcot, R. A. F. Romero, and V. Kumar (2020).
"Place Recognition in Forest with Urquhart Tessellations" in IEEE Robotics and Automation Letters (RA-L), 2020.
\vspace{5pt}

bibtex:

\vspace{5pt}

\begin{verbatim}
@inproceedings{nardari2020,
title={Place Recognition in Forest with Urquhart Tessellations},
author={Nardari, Guilherme V. and Cohen, Avraham and Chen, Steven W.
and Liu, Xu and Arcot, Vaibhav and Romero, Roseli A. F. and Kumar, Vijay},
booktitle={IEEE Robotics and Automation Letters (RA-L)},
year={2020}
}
\end{verbatim}

\twocolumn
\newpage

\title{Place Recognition in Forests with \\ Urquhart Tessellations}

\author{
    Guilherme V. Nardari$^{1,2}$, Avraham Cohen$^{2}$, Steven W. Chen$^{2,3}$,\\ Xu Liu$^{2}$, Vaibhav Arcot$^{2,3}$, Roseli A. F. Romero$^{1}$, and Vijay Kumar$^{2}$%
\thanks{
Manuscript received: September 6, 2020; Revised October 28, 2020; Accepted November 15, 2020.
This paper was recommended for publication by Editor Y. Choi upon evaluation of the Associate Editor and Reviewers' comments.
Work supported by Treeswift under the NSF SBIR grant \#193856, by ARL grant ARL DCIST CRA 2911NF-17-2-0181, ONR grant N00014-07-1-0829, ARO grant W911NF-13-1-0350, INCT-INSac grants CNPq 465755/2014-3, FAPESP 2014/50851-0 and 2018/24526-5. This work is supported in part by the Semiconductor Research Corporation and DARPA. (C}
\thanks{
$^{1}$G. V. Nardari and R. A. F. Romero are with the Robot learning laboratory, ICMC - University of Sao Paulo, Brazil. \texttt{\small (e-mail: guinardari@usp.br, rafrance@usp.br)} \newline
$^{2}$G. V. Nardari, A. Cohen, S. W. Chen, X. Liu, V. Arcot, and V. Kumar are with GRASP Laboratory, University of Pennsylvania, United States. \texttt{\small (e-mails: gnardari@seas.upenn.edu; avrahamc@seas.upenn.edu; chenste@seas.upenn.edu;
liuxu@seas.upenn.edu;
yvarcot@seas.upenn.edu;
kumar@seas.upenn.edu)} \newline
$^{3}$ S. Chen and V. Arcot are with Treeswift. \texttt{\small (e-mails steven@treeswift.com, vaibhav@treeswift.com)}}
}

\maketitle

\begin{abstract}
In this letter, we present a novel descriptor based on Urquhart tessellations derived from the position of trees in a forest. We propose a framework that uses these descriptors to detect previously seen observations and landmark correspondences, even with partial overlap and noise. We run ~\textcolor{black}{loop closure detection} experiments in simulation and real-world data map-merging from different flights of an Unmanned Aerial Vehicle (UAV) in a pine tree forest and show that our method outperforms state-of-the-art approaches in accuracy and robustness.
\end{abstract}


\section{Introduction}
    \label{sec:intro}
    \input{tex/intro}

\section{Problem Formulation}
    \label{sec:problem}
    \input{tex/problem}

\section{Preliminaries}
    \input{tex/definitions}

\section{Method}
    \label{sec:methodology}
    \input{tex/methodology}

\section{Experiments}
    \label{sec:exp}
    \input{tex/exp}

\section{Discussion and Conclusions}
    \label{sec:conc}
    \input{tex/conc}

\bibliographystyle{./bibliography/IEEEtran}
\bibliography{./bibliography/references}

\end{document}

%% file: tex/intro.tex
\IEEEPARstart{I}{dentifying} previously encountered locations is fundamental to a variety of robotic applications such as loop closure in simultaneous localization and mapping (SLAM) \cite{labbe2013appearance}, or merging observations from different robots in multi-robot systems \cite{labbe2014online}. 
This problem is more challenging in GPS-denied or restricted settings where absolute information about locations is unavailable or unreliable. Forests are an interesting and challenging use of robot systems, with promising applications to automatic timber volume estimation~\cite{chen2020sloam}, animal detection~\cite{van2014nature}, and forest fire management~\cite{merino2012unmanned}. The problem of place recognition is critical in cluttered forests as the operational area is vast, GPS is typically unreliable due to dense forest canopy, infrastructure for long-range communication is usually not available, and the environment is repetitive~\cite{tian2018search}.

\begin{figure}[t!]
\includegraphics[width=\columnwidth]{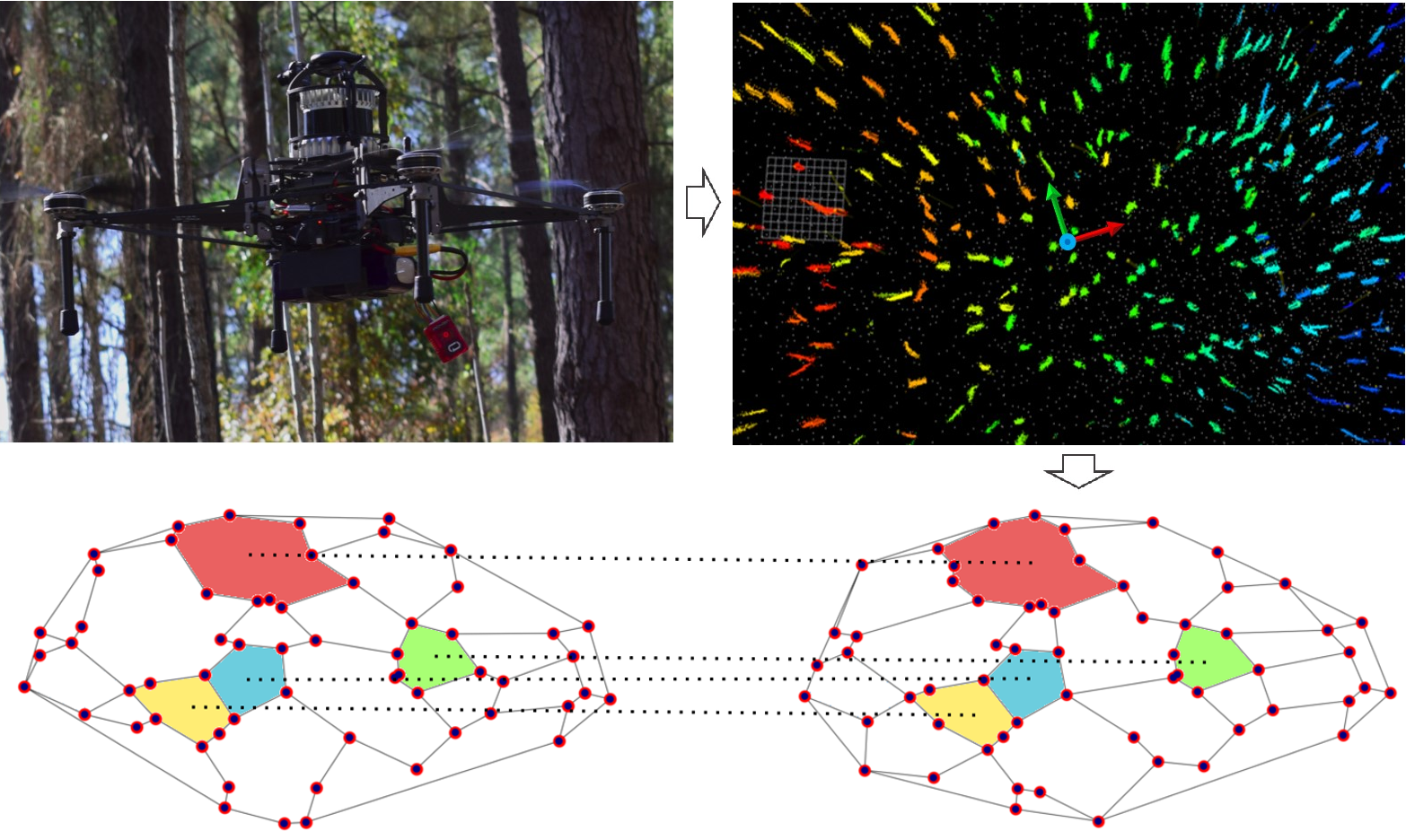}
\caption{As the robot detects landmarks, we can derive polygons based on their positions to detect loops and associate landmarks even if the observations are not completely consistent.}
\label{fig:robot_platform}
\vspace{-5mm}
\end{figure}

To address this, we propose a new method for place recognition illustrated in Fig.~\ref{fig:robot_platform} that derives polygons based on tessellations of the set of landmark detections. From these polygons, we extract unique descriptors of parts of a robot observation that can be used to localize the agent and compute landmark correspondences.

Many works address the place recognition problem in urban or indoor scenarios, where features have meaningful and discriminative information~\cite{galvez2012bags,lowry2015visual}. For the multi-view case, observations of an environment can be differentiated by encoding the semantic labels observed in a scene~\cite{gawel_x-view_2018}.
However, when navigating under the forest canopy, these methods are not viable since the only reliable information differentiating trees is their position in space. In this type of scenario, instead of a bag of words approach, one can resort to encoding the objects' spatial relationships using a global descriptor that encodes the entire observation.

Graphs are a natural model for these types of representations. The graph nodes are usually given by features or landmarks detected in an observation. A kernel can then be applied to extract a descriptor that can be matched against other observations~\cite{garg_lost?_2018}.
Methods such as GLARE~\cite{himstedt_large_2014}, its rotation invariant extension GLAROT~\cite{kallasi2016efficient} are similar in that they compute both a global descriptor for each observation, as well as local descriptors for each landmark in that observation based on the relative distances and angles between landmarks.
During the matching stage, they first check that the global descriptors are consistent, then associate landmarks based on the local descriptors.
Even though these methods have had some success in a forest environment~\cite{tian2018search}, they are susceptible to noise since they rely on space discretization to compute the local descriptors, and the estimate of the same landmark can lie in different bins.

For the 2-D case, Rizzini et al.~\cite{rizzini2019geometric} extends these works by representing the same metrics as probability density functions to avoid the discretization problem. However, this approach can still suffer from noise and fail to match observations that only partially overlap each other. We bypass these drawbacks through encoding local spatial relationships with geometric primitives, which is robust to partial overlaps and does not rely on discretization.

On the other end of the spectrum, local methods compute descriptors based on small regions around feature points. These methods can successfully match even in the presence of partial overlap or occlusion.
Gawel et al.~\cite{gawel2016structure} merge accumulated point cloud maps from vision and lidar systems by calculating structural descriptors based on the density of the neighborhood of feature points.
Fast point feature histograms (FPFH)~\cite{rusu2009fast} create descriptors for each point by computing surface normals using its neighboring points inside a user-defined radius. FPFH bins these normals into a histogram by representing them as angular features.

The drawback of local methods is that the number of comparisons usually grows linearly with the number of points in the set, resulting in higher computational times. For robotics applications where speed and resource allocation is critical, this can be prohibitively expensive.

Using polygons derived from the Urquhart graph, our method encodes information with fewer descriptors than a local method while also being robust to partial overlap and noise. We run experiments in simulation and a real-world forest, demonstrating that it can be used for loop-closure detection and landmark association for map-merging while handling noise perturbations that cause current state-of-the-art methods to fail. In summary, our contributions are:

\begin{itemize}
    \item A novel descriptor based on polygons capable of handling large scale repetitive environments while being robust to noise;
    \item A framework for aligning arbitrary observations using only the position information of landmarks.
\end{itemize}

%% file: tex/problem.tex
We model a robot with a noisy, limited range sensor traversing an environment filled with identical landmarks. Moreover, we define a map as a set of landmarks $\mathcal{M} \triangleq \{\mathcal{L}_{i}\}_{i=1}^{N}$, where the only information differentiating $\mathcal{L}_{i}$ and $\mathcal{L}_{j}$, $i\neq j$ are their locations in $\mathcal{M}$. Due to the limited range sensor, a robot at time $t$ has the potential to detect a submap of landmarks $\mathcal{S}^{M}(t) \subseteq \mathcal{M}$. However, due to sensor noise and landmark occlusion, some landmarks may not be detected. Let $\delta(\mathcal{L})$ be the Boolean random variable representing the successful detection of landmark $\mathcal{L}$,

\begin{equation*}
    \delta(\mathcal{L}) = 
    \begin{cases}
    1 & \text{if landmark }  \mathcal{L} \text{ is detected,}\\
    0              & \text{otherwise.}
    \end{cases}
\end{equation*}
Similarly to ~\cite{wang2011viewpoint}, we model $\delta(\mathcal{L})$ as a Bernoulli distribution with success probability $\omega$, $\delta(\mathcal{L}) \sim Ber(\omega)$. We define the observed submap under the presence of detection noise
\begin{equation*}
\bar{\mathcal{S}}^{M}(t) \triangleq \{\mathcal{L} : \delta(\mathcal{L})=1\}_{\mathcal{L} \in \mathcal{S}^{M}(t)} \subseteq \mathcal{S}^{M}(t).
\end{equation*}

We assume that each landmark $\mathcal{L}$ has a 2-D coordinate projection on the $xy$-plane in the map frame $\mathbf{p}^{M} \in \mathbb{R}^{2}$. Due to sensor noise and uncertainty in the landmark projection, the observed 2-D coordinate projection $\bar{\mathbf{p}}^{M}$ may differ from the true 2-D coordinate projection $\mathbf{p}^{M}$. We model this noise as $\bar{\mathbf{p}}^{M} = \mathbf{p}^{M} + \epsilon$, where $\epsilon$ is a 2-D Gaussian random variable with zero mean and variance $\Sigma$, $\epsilon \sim \mathcal{N}(\mathbf{0}, \Sigma)$. The 2-D observation of $\mathcal{S}^{M}(t)$ including both forms of sensor noise is 
\begin{equation*}
\mathcal{P}^{M}(t) \triangleq \{\bar{\mathbf{p}}^{M}\}_{\mathcal{L} \in \bar{\mathcal{S}}^{M}(t)}.
\end{equation*}

Finally, the robot will typically perceive its surroundings in its local frame, not the global map frame. Suppose that at time $t$, the pose of the robot projected in the $xy$-plane in the map frame is $\mathbf{T}^{M}_{R}(t) \in SE(2)$. We define the noisy 2-D observation of $\mathcal{S}^{M}(t)$ in this local frame as 
\begin{equation*}
\mathcal{P}^{R}(t) \triangleq \{[\mathbf{T}^{M}_{R}(t)]^{-1}\bar{\mathbf{p}}^{M}\}_{\mathcal{L} \in \bar{\mathcal{S}}^{M}(t)}.
\end{equation*}

We will refer to the robot pose at time $t$ as $\mathbf{T}(t)$ for ease of notation.

\textbf{Problem} (Place recognition with identical landmarks). Given noisy 2-D observations $\mathcal{P}^{R}(t_{i})$ and $\mathcal{P}^{R}(t_{j})$ taken at times $t_{i} \neq t_{j}$, determine if the corresponding sub-maps $\mathcal{S}^{M}(t_{i}) \cap \mathcal{S}^{M}(t_{j}) \neq \emptyset$ and if so, estimate the associated rigid transformation $\mathbf{H}(t_{i},t_{j}) \in SE(2)$ where

\begin{equation}
    \mathbf{T}(t_{i}) = \mathbf{H}(t_{i},t_{j})\mathbf{T}(t_{j}).
    \label{eq:problem_tf}
\end{equation}

This problem encapsulates a variety of different applications such as kidnapped robot~\cite{engelson1992error}, multi-robot map fusion~\cite{tian2018search}, and loop closure detection~\cite{pierzchala2018mapping} in environments with identical landmarks (e.g. forests).

%% file: tex/definitions.tex
\label{sec:definitions}

An edge $e = (p_i, p_j)$ is a line segment bounded by a pair of points $p_i$, $p_j \in \mathbb{R}^2$. A polygon $L$ is a closed set defined by the region enclosed by the edges constructed via consecutive point pairs $(p_i, p_{i+1})$ in the sequence of points $(p_0, \dots, p_n)$ where $p_0 = p_n$. In other words, polygon $L$ is defined by a sequence of edges $(e_0, \dots, e_{n-1})$, where each edge $e_i$ is constructed based on the original point sequence. A triangle is a polygon with $n = 3$, and its circumcircle is the circle that passes through all points of the triangle.

Let $P$ be a set of at least $3$ discrete points $p \in \mathbb{R}^{2}$ in general position.
 A tessellation is a finite set of polygons $\{L_{0},\dots L_{n}\}$ which covers the convex hull $\mathcal{Q}(P)$ without gaps or overlaps. More precisely, $\cup_{i=0}^{n} L_i = \mathcal{Q}(P)$ and $int(L_i) \cap int(L_j) = \emptyset \, ~\forall \, i \neq j$, where $int(\cdot)$ denotes the interior of a polygon. A triangulation is a tessellation where all elements are triangles. The Delaunay triangulation $DT(P)$ is a triangulation where no point $p \in P$ is in the circumcircle of any triangle of $DT(P)$~\cite{delaunay1934sphere}. 

We convert the Delaunay triangulation to a graph in order to construct the Urquhart graph~\cite{urquhart1980algorithms}. The triangulation $DT(P)$ can be represented as a graph $\mathcal{G_{D}} = \{\mathcal{V_{D}}, \mathcal{E_{D}}\}$ where $\mathcal{V_{D}}$ is the union of the triangle points, and $\mathcal{E_{D}}$ is the union of the triangle edges. The set of the longest edges of each triangle in $DT(P)$ is defined by $\Omega = \{\argmax_{e \in L} \left \| e \right \| : L \in DT(P)\}$. The Urquhart graph of $DT(P)$ is a graph $\mathcal{G}_{U} = \{\mathcal{V}_{U}, \mathcal{E}_{U}\}$ where $\mathcal{V}_{U} = \mathcal{V_{D}}$ and $\mathcal{E}_{U} = \mathcal{E_{D}} \setminus \Omega$. $\mathcal{G}_{U}$ is a sub-graph of $\mathcal{G_{D}}$ where the longest edges of each triangle are removed.

We then convert the Urquhart graph $\mathcal{G}_{U}$ back into an Urquhart tessellation $U(P)$ using cycle detection. A simple cycle $c$ of an arbitrary graph $\mathcal{G} = \{\mathcal{V}, \mathcal{E}\}$ is a non-empty sequence of edges $\mathcal{E}_{c} = (e_0,\dots,e_{n-1}) \subseteq \mathcal{E}$ with a vertex sequence $\mathcal{V}_{c} = (v_0,\dots,v_n) \subseteq \mathcal{V}$ such that $v_0 = v_n$, and $v_i = v_j \iff$  $i,j \in \{0, n\}$, i.e. there are no repeated vertices except for the first and the last. The simple cycles of a graph correspond to polygons of a tessellation.

The cycle basis $\mathcal{C}$ of a graph $\mathcal{G}$ is the minimal set of simple cycles such that for all cycles $c \in \mathcal{G}$, $\exists \, c_i,c_j \in \mathcal{C}$, such that $c = c_i \Delta c_j$ where $\Delta$ represents the symmetric difference operation. Intuitively, any cycle can be computed with elements of the cycle basis. The cycle basis of a graph corresponds to the tessellation. As a result, we can convert the graph $\mathcal{G}_{U}$ into the tessellation $U(P)$.

These tessellations motivate a hierarchy of geometric primitives $\mathcal{H}(P)$ that encompass local to global information. The first level $\mathcal{H}^0(P)$ is the set of all edges $\mathcal{E}_D$. The second level $\mathcal{H}^1(P)$ is given by the triangles of the Delaunay triangulation $DT(P)$. The third level $\mathcal{H}^2(P)$ is given by the polygons of the Urquhart tessellation $U(P)$.

We define a function $\phi^{k}(\cdot): \mathcal{H}^{k+1}(P) \rightarrow \mathcal{H}^{k}(P)$, to map from higher to lower levels, where 
\begin{equation}
     \phi^k(s) = \{l: l \cap s = l, l \in \mathcal{H}^{k}(P)\},
     \label{eq:phi}
\end{equation}

e.g. $\phi^0$ maps triangles of the Delaunay triangulation to its corresponding edges, $\phi^1$ maps polygons of the Urquhart tessellation to its corresponding triangles.

%% file: tex/methodology.tex
\begin{figure*}
    \centering
    \includegraphics[width=0.9\textwidth]{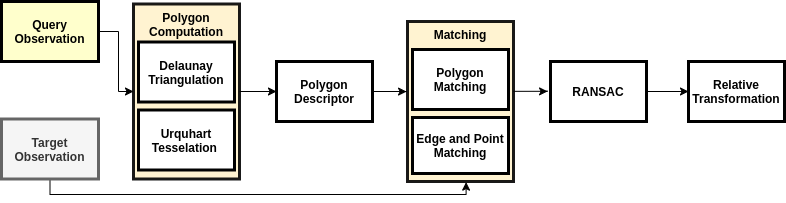}
    \caption{Our method computes tessellations from the position of trees detected from lidar scans. The polygons derived from it can be used for place recognition and landmark association. The target observation is a different set of polygons with descriptors that can come from the robot's history of observations in loop closure tasks or a different agent in multi-robot scenarios.}
    \label{fig:pipeline}
    \vspace{-5mm}
\end{figure*}

In this section, we describe our method for extracting from an observation $\mathcal{P}^{R}(t)$, the polygons $\mathcal{H}(\mathcal{P}^{R}(t))$ or $\mathcal{H}(t)$ for brevity. Moreover, we present a framework depicted in Fig.~\ref{fig:pipeline} for solving Eq.~\ref{eq:problem_tf} using landmark correspondences derived from $\mathcal{H}(t)$. We assume that the spatial distribution of the landmarks is unique, and leverage these landmarks' positions to define polygons that can be used to represent a sub-map $\mathcal{S}(t)$ uniquely.

\textcolor{black}{Our method expects a set of noisy 2-D landmark locations as input. Li et al.~\cite{li2020localization} obtains this information using a clustering algorithm.}
\textcolor{black}{We derive the position of the trees observed by the robot with SLOAM, a framework for odometry and mapping in forests proposed in our previous work~\cite{chen2020sloam}. The algorithm detects tree instances in lidar point clouds and estimates a cylinder model for each tree. Each cylinder is parameterized by a ray that defines its axis and a radius. The position of the trees is estimated by projecting the initial point of the ray onto the 2-D plane.} \textcolor{black}{Any method that can output consistent landmark locations relative to the robot across observations can be used for this step. The choice depends on the type of robot or environment.}

\subsection{Polygons from Landmark Detections}
\label{ssec:polys}
In general, a triangle of $DT(t)$ will have consistent metric properties such as perimeter and area across observations as long as other polygons that share a side with it are not perturbed with noise~\cite{li2020localization}. However, similar triangles are likely to be found as the scale of the area covered by the robot grows. Moreover, the number of triangles to match has an upper bound of $2n-2-b$ where $n$ is the number of points and $b$ are the points that lie in $\mathcal{Q}(P)$~\cite{delaunay1934sphere}, which can become impracticable to match in real-time applications.

For two polygons $L_m$ and $L_n$, $L_m \cap L_n = \emptyset$, from different regions of the map to have similar metric properties, it would require that the triangles that compose $L_m$ and $L_n$ also have similar metric properties and are arranged in space similarly, such that $\bigcup \, \phi^{1}(L_m) \approx \bigcup \, \phi^{1}(L_n)$. For this reason, $U(P)$ creates polygons that are less likely to repeat than $DT(P)$ and decreases the probability of false-positive correspondences.

As stated in Sec.~\ref{sec:definitions}, the set of polygons of the tessellation $U(t)$ is given by the cycle basis $\mathcal{C}_{U}$. We propose an algorithm summarized in ~\textcolor{black}{Alg.}~\ref{alg:urqu_cycle_detection}, where we loop through the elements of $\mathcal{H}^1(t)$ to compute the set of longest edges $\Omega$ while also updating $\mathcal{C}_{U}$ as triangles are combined to efficiently compute both $\mathcal{G}_U$ and $\mathcal{C}_U$.

\begin{figure}
\centering
\includegraphics[width=0.6\columnwidth]{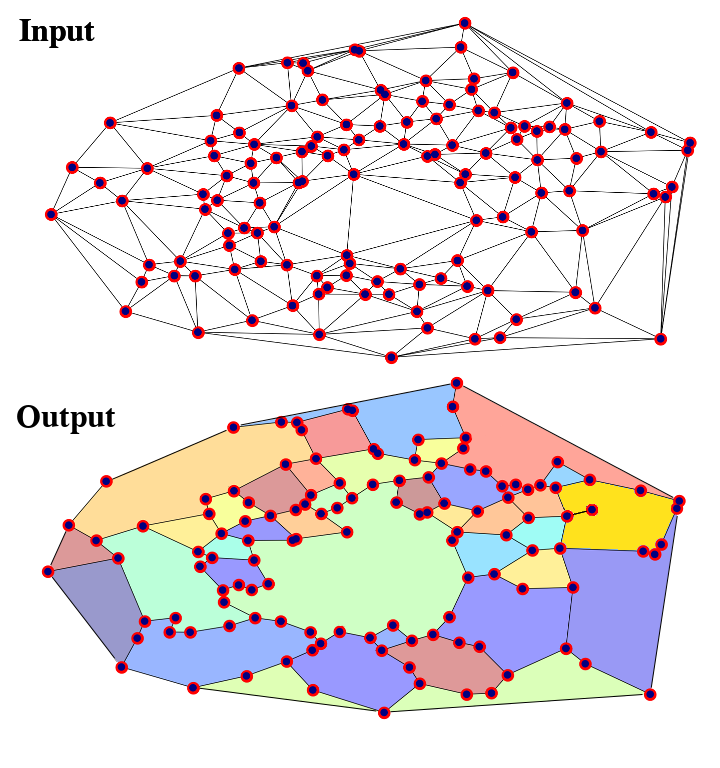}
    \caption{A graph before and after Algorithm \ref{alg:urqu_cycle_detection}. It drops the longest edge for each triangle while keeping track of the cycles being formed (represented by the colored regions).}
    \label{fig:urquhart}
    \vspace{-15pt}
\end{figure}

For a given pair of cycles $c_a$ and $c_b$ that represent polygons from a tessellation, we compute the symmetric difference $c_a \Delta c_b$ by moving the elements of its lists of edges $\mathcal{E}_{c_a}$ and $\mathcal{E}_{c_b}$. The longest edge of $c_a$ that is shared with $c_b$, $e \in \mathcal{E}_{c_a}, e \in \mathcal{E}_{c_b}$ is moved to the first position in $\mathcal{E}_{c_b}$ and the last position in $\mathcal{E}_{c_a}$. Then, we concatenate $\mathcal{E}_{c_b}$ to the end of $\mathcal{E}_{c_a}$, excluding $e$ from both. This procedure will create a new cycle and maintain the order of the edges. In some cases, elements of $\mathcal{C}_U$ will have a hanging edge, as depicted by the yellow polygon in ~\textcolor{black}{Fig.}~\ref{fig:urquhart}. These elements can be detected and filtered out in post-processing to respect the definition of a simple cycle. Furthermore, polygons that have a side at the boundary of the tessellation are discarded since these shapes have a probability of changing in a new observation with different robot position as new landmarks can enter the field of view of the sensor.

\begin{algorithm}[t!]
	\caption{Urquhart Graph with Cycle Detection}
	\begin{algorithmic}[1]
    \State \textbf{input}:  $\mathcal{G}_{D}$, $\mathcal{H}^1(t)$
    \State $\mathcal{C}$ = $\mathcal{H}^1(t)$
    \State $\mathcal{G}_{U}$ = $\mathcal{G}_{D}$
     \For{each triangle $L \in \mathcal{H}^1(t)$}
        \State $e_L = \argmax_{e \in L} \left \| e \right \|$
        \State Find $L_{neigh} \in \mathcal{C}$, $L_{neigh} \neq L$, $e_L \in L_{neigh}$ \Comment{Find a neighboring triangle that shares $e_L$.}
        \State Drop $e_L$ from $\mathcal{G}_{U}$
        \State $\mathcal{C}_L = L \Delta L_{neigh}$
    \EndFor
    \State \Return  $\mathcal{G}_{U}$, $\mathcal{C}$
	\end{algorithmic}
	 \label{alg:urqu_cycle_detection}
\end{algorithm}

\subsection{Robust Polygon Descriptor}
We can view the place recognition problem as an instance of the sub-graph matching problem using $\mathcal{G}_{U}$, which is NP-Complete~\cite{cook1971complexity}. Instead, we use $\mathcal{H}(t)$ to derive descriptors for different regions of the observation.
~\textcolor{black}{To compute our descriptor, we assume that observations will not be sheared or have scale differences, which will be true for lidar data.}

We borrow techniques from the shape retrieval literature and, for each polygon and triangle $L \in \mathcal{H}(t)$,~\textcolor{black}{we compute a descriptor based on the centroid distance~\cite{zhang2001comparative}}. The set of points that are part of $L$ is $N = \{p : p \in L\}$ where $p$ are the elements of the sequence of points from the definition of a polygon in Sec.~\ref{sec:definitions}, excluding the repeated point. The centroid $c = (c^x, c^{y})$ of $L$ is computed by

\begin{equation*}
    c^x = \dfrac{1}{\left | N \right |} \sum_{n=0}^{\left | N \right |-1} p_{n}^{x}, \quad c^y = \dfrac{1}{\left | N \right |} \sum_{n=0}^{\left | N \right |-1} p_{n}^{y}.
\end{equation*}

Since the size of $N$ can vary for different polygons, we sample a constant number points relative to the perimeter size. The step size between sampled points is given by $step * P$, where $0 < step < 1.0$ and $P$ is the length of the perimeter, creating a new set of points $M$ with the same number of elements regardless of the size of $N$.

The new \textcolor{black}{centroid distance} with sampled points is
\begin{equation*}
    F(L) = \{ \left \| p_n - c \right \|^2 \, : \, p_n \in M \}.
\end{equation*}

A large $step$ size will smooth out the polygon, while a small one will be more likely to capture details such as sharp corners. The optimal value balances these two properties to maximize precision while also being robust to noise.

$F(L)$ is translation invariant since it uses only relative distances. However, the order of the elements of the descriptor can be different depending on what part of the polygon sampling starts. Similar to GLAROT~\cite{kallasi2016efficient}, one could apply a permutation function to the descriptor and use the configuration with the smallest distance. This can be inefficient, as we may have to match many polygons per observation.

We address this problem by applying a Discrete Fourier Transform $DFT(\cdot)$ to $F(L)$, and obtain a new descriptor $\hat{F}(L) = DFT(F(L))$~\cite{zhang2001comparative}. $\hat{F}(L)$ is in the frequency domain, and has the property that its magnitude $\bar{F}(L) = \left | \hat{F}(L) \right |$ will be the same regardless of the order of the input, making it invariant to starting point of the sampling step.

\subsection{Matching}
We store $\bar{F}(L)$ for all polygons in $\mathcal{H}(t)$, including triangles.
A pair of polygons $L_n \in \mathcal{H}(t_i)$ and $L_m \in \mathcal{H}(t_j)$ are considered a match if $\left \| \bar{F}(L_n) - \bar{F}(L_m) \right \|^2 <  \tau$. To increase robustness and speed, we only compare polygons if $\left | N_n \right | - \left | N_m \right | \leq 3$. That is, if the difference in number of points between polygons is smaller than or equal to $3$. 

In the worst case, the initial comparison between elements of $\mathcal{H}^{2}(P)$ of a pair of observations will be $g_{t_i} * g_{t_j}$ where $g_i =  \left | \mathcal{H}^{2}(t_i) \right |$ and $g_j =  \left | \mathcal{H}^{2}(t_j) \right |$. In loop-closure or other scenarios where the number of comparisons scales over time with the number of stored observations, we may have to sample elements of $\mathcal{H}^{2}(t)$.

We define a parameter $\gamma$ that creates an upper bound on the number of $\mathcal{H}^{2}(t)$ polygons per observation and, consequently, the time required for matching a pair of observations. 
With this process, the algorithm will still grow linearly with respect to the number of stored observations. However, like in global methods, the number of landmarks will no longer affect the performance since the number of polygons to match per observation will be constant.

As stated previously, a polygon $L_m \in \mathcal{H}^{2}(t_i)$ defines a subset of triangles $\phi^1(L_m) \subseteq \mathcal{H}^{1}(t_i)$. Given a target polygon $L_n  \in \mathcal{H}^{2}(t_j)$, if the ratio between elements in $\phi^1(L_m)$ with correspondences in $\phi^1(L_n)$, and the number of elements of $\phi^1(L_m)$ is greater than a threshold $eta$, the triangle correspondences are considered valid.

For a pair of corresponding triangles $L_k \in \mathcal{H}^{1}(t_i)$ and $L_l \in \mathcal{H}^{1}(t_j)$, we match edges based on their lengths $\bar{\bar{\mathcal{E}_k}} = \{\left \| e \right \| : e \in L_k\}$, and $\bar{\bar{\mathcal{E}_l}} = \{\left \| e \right \| : e \in L_l\}$ by

\begin{equation*}
    \argmin_k \left \| \bar{\bar{\mathcal{E}_m}} - \chi_k  \bar{\bar{\mathcal{E}_n}} \right \|^2,
\end{equation*}

where $\chi$ is a permutation matrix reordering the elements of $\bar{\bar{\mathcal{E}}}$. Intuitively, the matrix that generates the smallest difference between the lengths of the edges is the best assignment between them. We extend this assignment to point correspondences by matching points that share corresponding edges.

\subsection{Euclidean Transformation Computation}
\label{ssec:ransac}
Given a set of correspondences, there are many approaches to solving the observation alignment problem. For the $\mathbb{R}^3$ case, the assignments found by our algorithm can be propagated to the entire object, and an optimization-based approach can be used to align the instances, e.g., as an alternative to the data association methods presented in~\cite{chen2020sloam} Sec. III-B.

\textcolor{black}{We run our experiments in $\mathbb{R}^2$, and use the assumption that the data will not suffer from shearing or scale variations to reduce $\mathbf{H}(t_{i},t_{j})$ to a Euclidean transformation with $3$ degrees of freedom that we estimate using RANSAC.}

For each iteration, we randomly sample two correspondences, and solve Eq.~\ref{eq:problem_tf} for $\mathbf{H}(t_{i},t_{j})$. If the Euclidean distance between corresponding points after the transformation is below a threshold $d$, we consider the correspondence an inlier. If the ratio of inliers to outliers is above a threshold $r$ or the maximum number of iterations $s$ is reached, the algorithm stops and returns the best estimate which has the most inliers.

%% file: tex/exp.tex
In this section, we present our experimental setup, analyze the influence of key parameters on our method's performance and compare it to other algorithms on loop closure detection and map-merging tasks.

\subsection{Setup}
We run our experiments in two different environments. The first is a simulated forest. We use ground truth pose measurements to evaluate our algorithm's precision and recall under varying noise levels and how different parameter configurations affect performance. In the second experiment, ~\textcolor{black}{we collected a real-world dataset at a commercial pine tree forest, manually flying a UAV under the tree canopy. As shown in Fig.~\ref{fig:robot_platform}, the UAV carries an Ouster OS1-64 $360^o$ 3D lidar, which has a $45^o$ field of view and $64$x$2048$ of vertical and horizontal resolution, respectively. The robot has an Intel® NUC 8i7BEK computer onboard to store the lidar data. We evaluate the quality of the associations detected by our algorithm in this dataset with a map merging task.}

We compare our method against GLARE with GLAROT distance~\cite{kallasi2016efficient}. The angle and distance discretization resolutions are $8$ and $50$. Following the original paper, we compute descriptors for every landmark and a global descriptor. If the GLAROT distance between two global descriptors is under a threshold, they are considered a match. Then, the landmark \textcolor{black}{descriptors} are used to detect associations. We also compare with ~\textcolor{black}{Li et at.~\cite{li2020localization}}. Developed in parallel with this work, they propose a landmark association method based on Delaunay triangulations. Their method computes a descriptor for each triangle by concatenating the areas and perimeters of itself and other triangles that share a side with it.

For our method, $\eta$ is set at $0.5$ in all experiments. For all methods, we use RANSAC to compute $\mathbf{H}(t_{i},t_{j})$ according to Sec.~\ref{ssec:ransac} with $d = 0.5$, $r = 0.99$ and $s = 40000$. Since the original papers do not explicitly recommend values for distance thresholds, we use the best configuration found empirically for GLAROT~\cite{kallasi2016efficient} and for Li et al.~\cite{li2020localization} which are $10$ and $100$ respectively.

\subsection{\textcolor{black}{Loop Closure Detection in Simulation}}
\begin{figure*}
    \begin{subfigure}{\columnwidth}
        \includegraphics[width=\columnwidth]{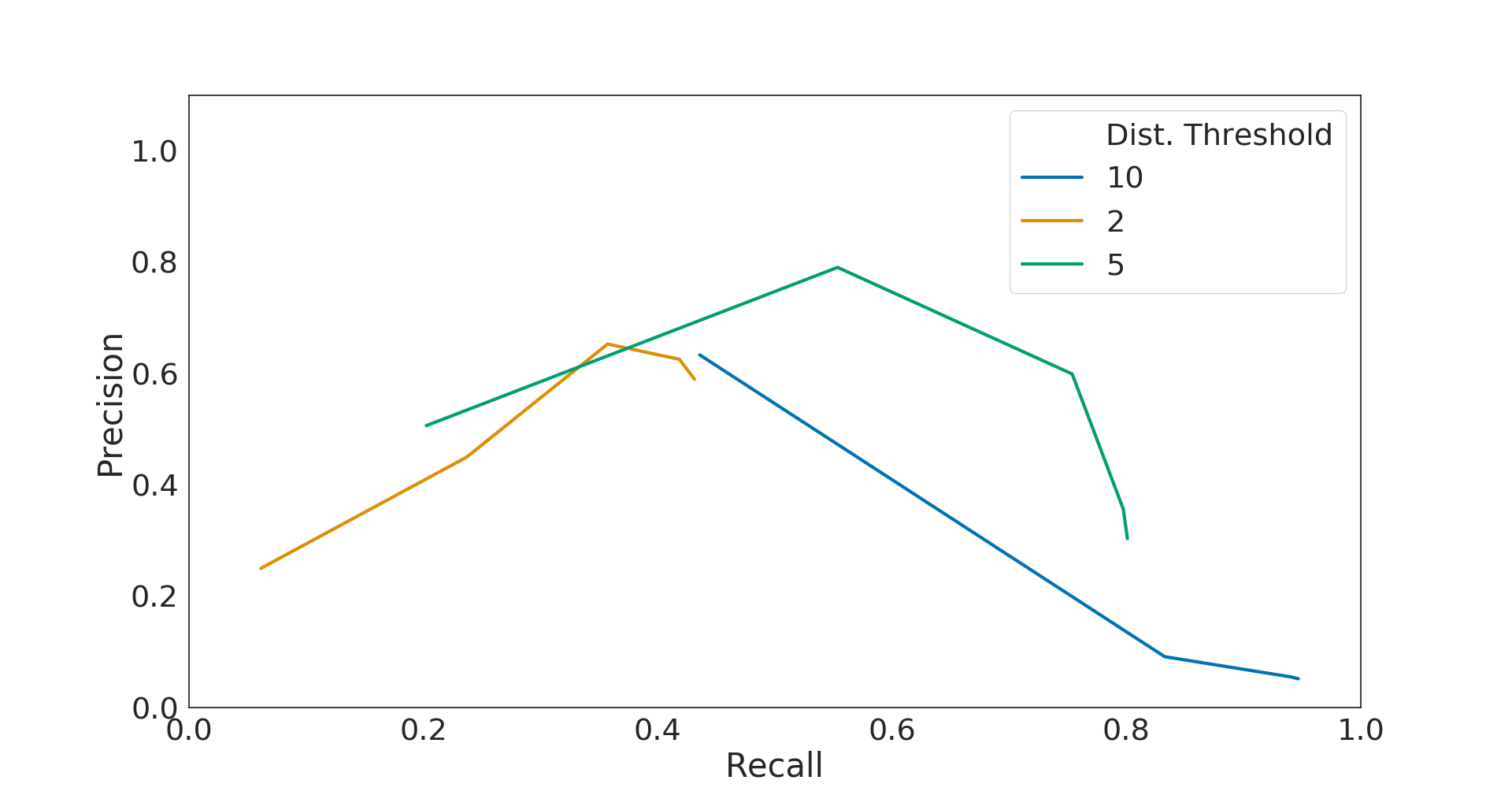}
        \caption{}
        \label{sfig:pr_poly_dist_thresh}
    \end{subfigure}
    ~
    \begin{subfigure}{\columnwidth}
        \includegraphics[width=\columnwidth]{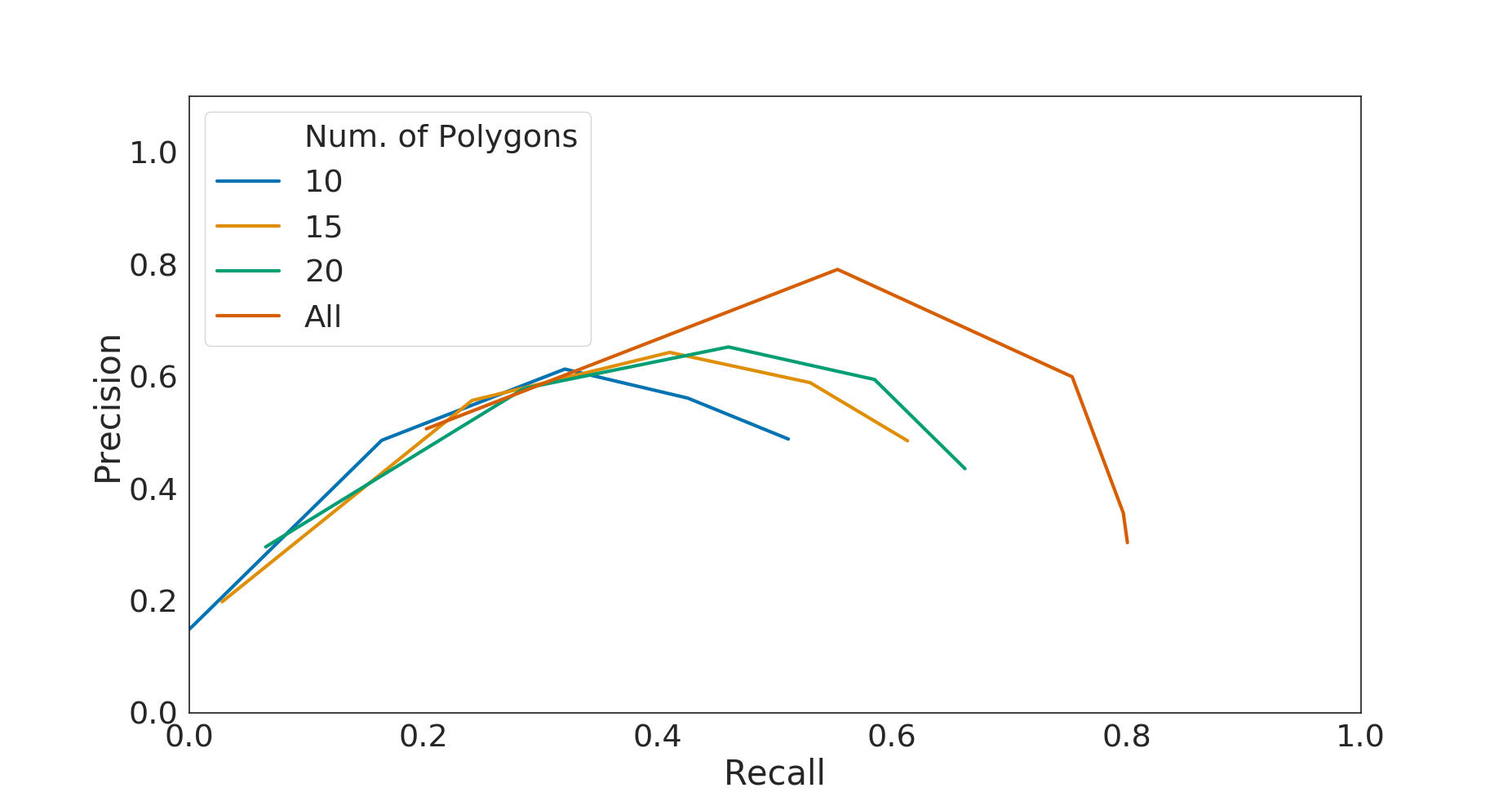}
        \caption{}
        \label{sfig:pr_poly_sample}
    \end{subfigure}
    \caption{(a) PR-curves for different distance thresholds. Our method achieves the best F1-score with $\tau = 5$. (b) PR-curves for different numbers of sampled polygons of $\mathcal{H}^2$, $\gamma$. We vary the number of point correspondences required to compute the transformation between observations to compute the curve for both experiments.}
    \label{fig:prcurves}
    \vspace{-5mm}
\end{figure*}

\textcolor{black}{In repetitive environments such as a forest, the challenging part of performing loop closure is detecting previously seen locations. Once this is solved, methods such as Factor graphs~\cite{dellaert2012factor} can be employed to incorporate the detection. For this reason, in this experiment, we focus on the sub-task of loop closure detection to measure the robustness of each method.}

We simulate a $1 km^2$ or approximately $247$ acres forest. To ensure a consistent density of trees across the map, the set of 2-D landmarks is generated by Poisson-Disc sampling through Bridson's algorithm~\cite{bridson2007fast} with a minimum distance between points of $7m$. This algorithm will create a regular pattern of trees across the environment, which is not a realistic representation of the distribution of trees in a real forest. To account for this, each point in the set is perturbed with Gaussian noise with $0$ mean and $3m$ standard deviation. After this steps, each $50m$ radius observation will have approximately $80$ trees per lidar reading and the average distance from a tree to its nearest neighbor is $3.4m$, while in the accumulated map from our real-world dataset this distance is $3.2m$.

The simulated robot does a circular path $4$ times. Each observation is rotated by a random angle sampled from a uniform distribution in the range $\{0, \dfrac{\pi}{2}\}$, and every landmark $\mathcal{L}$ in the observations subject to position noise $\epsilon$ and detection noise $\delta(\mathcal{L})$. Excluding the trivial match where $i = j$, we consider a match a true positive if $\left \| \mathbf{T}(t_{i}) - \mathbf{H}(t_{i},t_{j})\mathbf{T}(t_{j}) \right \|^2 < 10$ and the rotation difference is smaller than $20^o$, which are similar constrains to related work~\cite{gawel_x-view_2018}. We consider a false negative when not enough matches are found but the distance between the ground truth poses is smaller than the lidar radius.

For all experiments we run all possible combinations of the detection success probability $\omega$ in the range $\{0.8, 0.9, 0.95, 1.0\}$ and the standard deviation $\sigma = \sqrt{\Sigma}$ of the position estimation error $\epsilon$ in the range $\{0.0, 0.1, 0.2, 0.3, 0.4\}$ totalling 20 different experiments.

We compute Precision-Recall (PR) curves with respect to the minimum number of point to point associations in the range $\{4,8,16,32,64\}$ required to run RANSAC. For our method, we first store all polygons in $\mathcal{H}(t)$ and evaluate different configurations of the polygon descriptor distance $\tau$. The values of the PR curve are the average precision and recall under different combinations of values of $\epsilon$ and $\delta$. We consider the best configuration with respect to the F1-score, given by $2* \frac{(precision * recall)} {(precision + recall)}$.

By running the simulation experiment with different configurations of $\tau$, we find that with $\tau = 5$, our method achieves the best F1-score as shown in Fig.~\ref{sfig:pr_poly_dist_thresh}.

With this result, we evaluate the effect of the parameter $\gamma$. We select polygons of $\mathcal{H}^{2}(t)$ by prioritizing elements with $4 \leq N \leq 9$ since polygons with large $N$ are usually more sensitive to noise, and triangles encode less information about the space. If not enough polygons in this range are found, we continue to select randomly from the available polygons with $N > 9$ and, if still necessary, from triangles.

We show the PR curves with different values of $\gamma$ in Fig.~\ref{sfig:pr_poly_sample} and observe that sampling from elements from  $\mathcal{H}^{2}(t)$ has a direct impact on performance. For this reason, we use all polygons in the next experiments. 

Finally, we combine the best results of previous experiments to evaluate the effect of different noise levels on the precision and recall of each method. In \textcolor{black}{Table}~\ref{tab:sim_results} we show the performance of all methods for each combination of noise using the same range for $\sigma$ and $\delta$ as previous experiments.

Increasing the position noise or decreasing the number of detections impacts all methods' performance. Li et al.~\cite{li2020localization} can handle more detection failures than both our method and GLAROT. As triangles capture a smaller portion of the observation, they are more likely to have consistent polygons, even with a large percentage of unobserved landmarks. However, with $10cm$ of position noise, the performance of Li et al.~\cite{li2020localization} significantly drops. The main factor for this is that the descriptor relies on the area of the triangles, which has high sensitivity to noise.

For all methods, we used the configuration that achieved the best F1-score for our experiments. However, for GLAROT, that implied in either high precision and low recall or low precision and high recall. In this experiment, we observe many false positives, making GLAROT have poor performance even on the scenario with no noise. 

For our method, as the polygons in $\mathcal{H}^{2}(t)$ capture a larger area of the observation, these elements are more likely to be altered as landmarks are not detected. For this reason, while it is more robust than the other methods, we observe that our approach is more sensitive to detection failures.

\begin{table*}
\centering
\caption{F1-scores for each method in the simulation experiment. We simulate the observations of a robot with different combinations of landmark position noise and detection success probability.}
\resizebox{\textwidth}{!}{%
  \begin{tabular}{|l|ccc|ccc|ccc|ccc|}
    \hline
    \multirow{2}{*}{\backslashbox{Position Noise}{Detection Success Prob.}} &
      \multicolumn{3}{c|}{$100\%$} &
      \multicolumn{3}{c|}{$95\%$} &
      \multicolumn{3}{c|}{$90\%$} &
      \multicolumn{3}{c|}{$80\%$} \\
    & Ours & GLAROT~\cite{kallasi2016efficient} & Li et al.~\cite{li2020localization} & Ours & GLAROT~\cite{kallasi2016efficient} & Li et al.~\cite{li2020localization} & Ours & GLAROT~\cite{kallasi2016efficient} & Li et al.~\cite{li2020localization} & Ours & GLAROT~\cite{kallasi2016efficient} & Li et al.~\cite{li2020localization}  \\
    \hline
    0cm & $\mathbf{1.00}$ & $0.52$ & $\mathbf{1.00}$ & $\mathbf{1.00}$ & $0.27$ & $\mathbf{1.00}$ & $0.95$ & $0.12$ & $\mathbf{0.99}$ & $0.32$ & $0.01$ & $\mathbf{0.75}$  \\
    \hline
    10cm & $\mathbf{1.00}$ & $0.39$ & $0.07$ & $\mathbf{0.99}$ & $0.12$ & $0.01$ & $\mathbf{0.92}$ & $0.04$ & $0.00$ & $0.32$ & $0.01$ & $0.00$  \\
    \hline
    20cm & $\mathbf{0.99}$ & $0.17$ & $0.00$ & $\mathbf{0.98}$ & $0.06$ & $0.00$ & $\mathbf{0.82}$ & $0.02$ & $0.00$ & $0.23$ & $0.00$ & $0.00$  \\
    \hline
    30cm & $\mathbf{0.97}$ & $0.04$ & $0.00$ & $\mathbf{0.76}$ & $0.01$ & $0.00$ & $0.45$ & $0.00$ & $0.00$ & $0.08$ & $0.00$ & $0.00$  \\
    \hline
    40cm & $0.66$ & $0.01$ & $0.00$ & $0.30$ & $0.00$ & $0.00$ & $0.12$ & $0.00$ & $0.00$ & $0.01$ & $0.00$ & $0.00$  \\
    \hline
\end{tabular}%
}
\label{tab:sim_results}
\vspace{-2mm}
\end{table*}

\subsection{Map Merging}
\begin{figure*}
    \begin{subfigure}{0.23\textwidth}
        \includegraphics[width=\columnwidth]{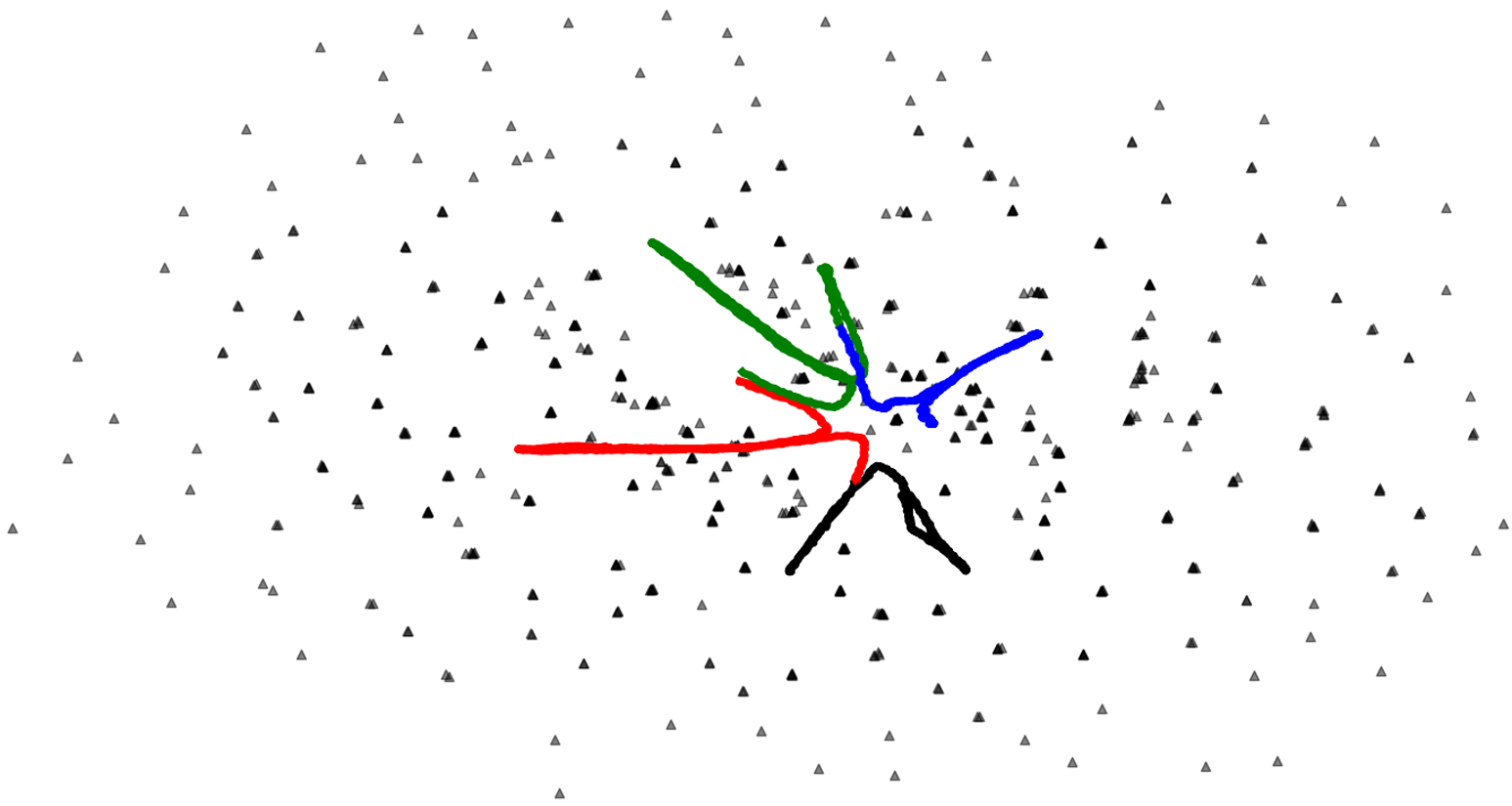}
    \end{subfigure}
    ~
    \begin{subfigure}{0.23\textwidth}
        \includegraphics[width=\columnwidth]{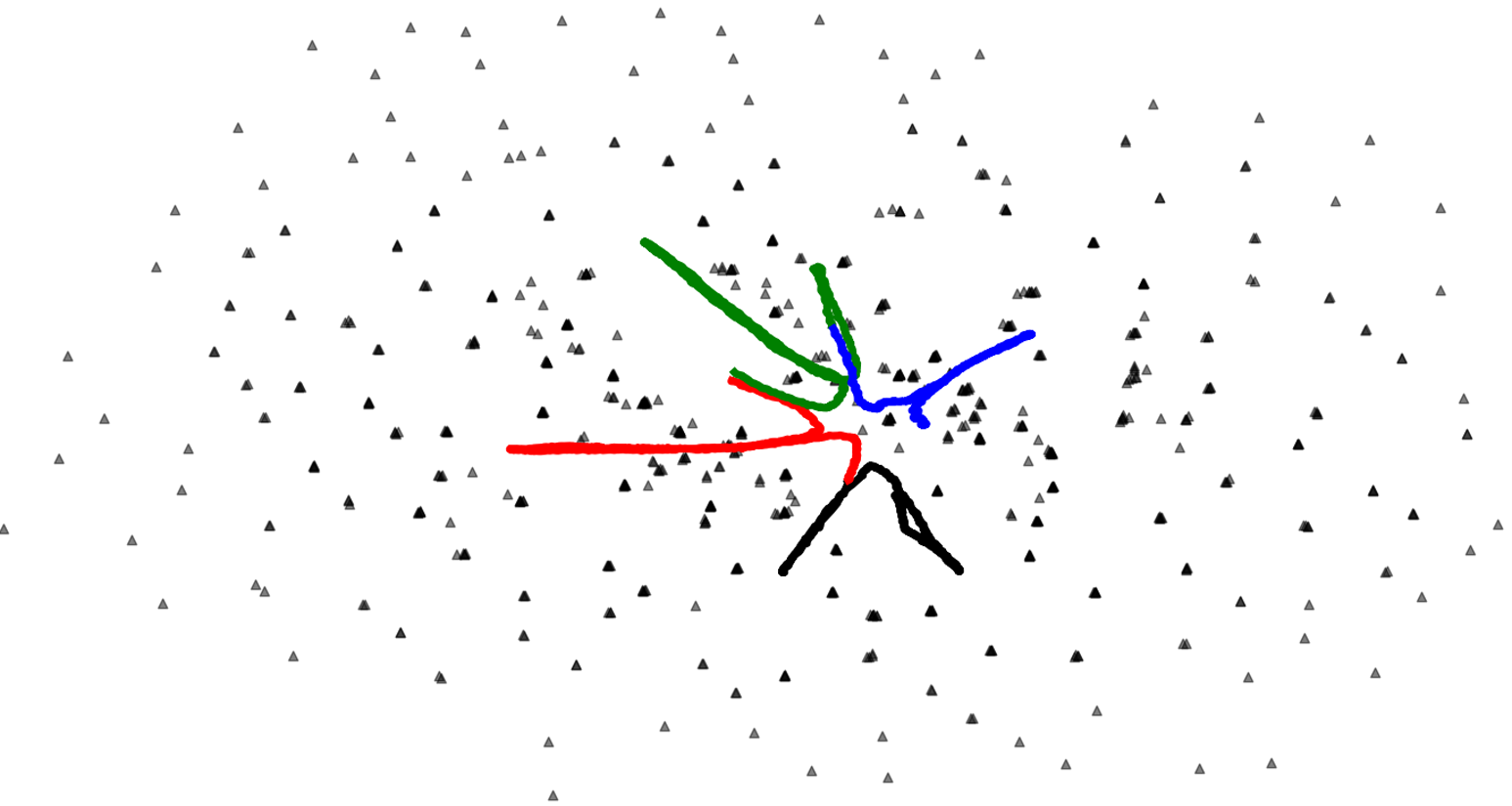}
    \end{subfigure}
    ~
    \begin{subfigure}{0.23\textwidth}
        \includegraphics[width=\columnwidth]{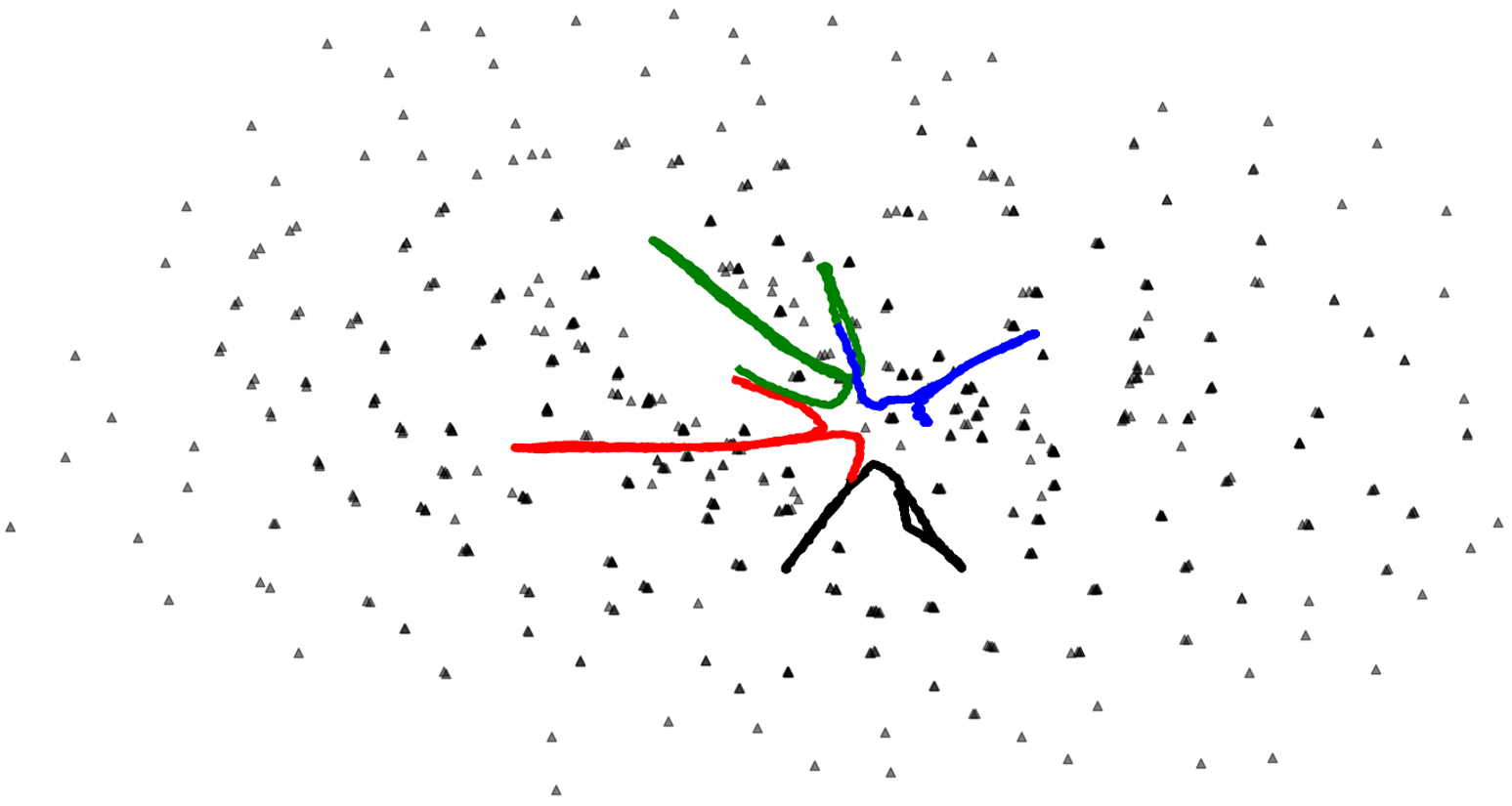}
    \end{subfigure}
    ~
    \begin{subfigure}{0.23\textwidth}
        \includegraphics[width=\columnwidth]{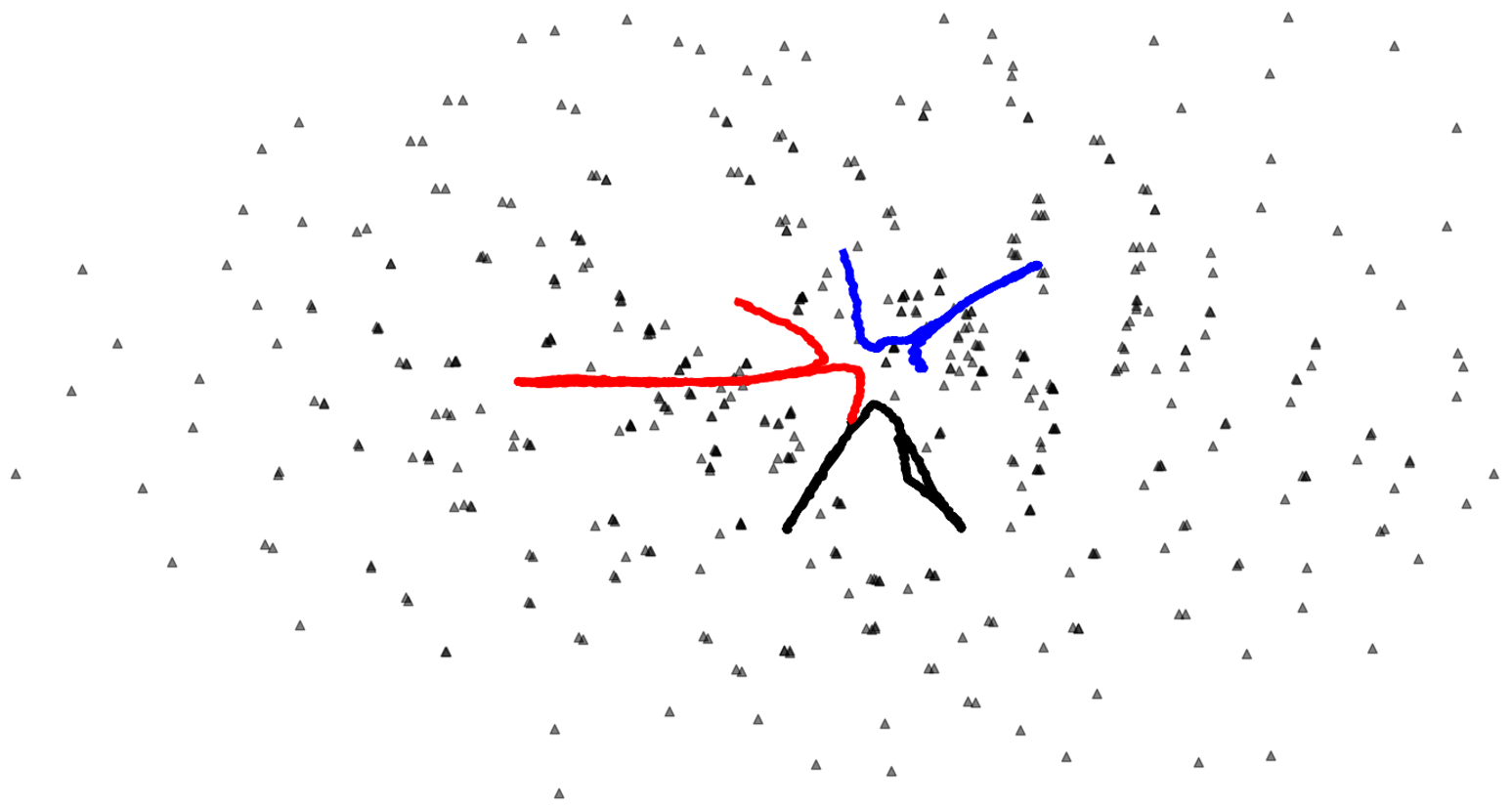}
    \end{subfigure}
    ~
    \begin{subfigure}{0.23\textwidth}
        \includegraphics[width=\columnwidth]{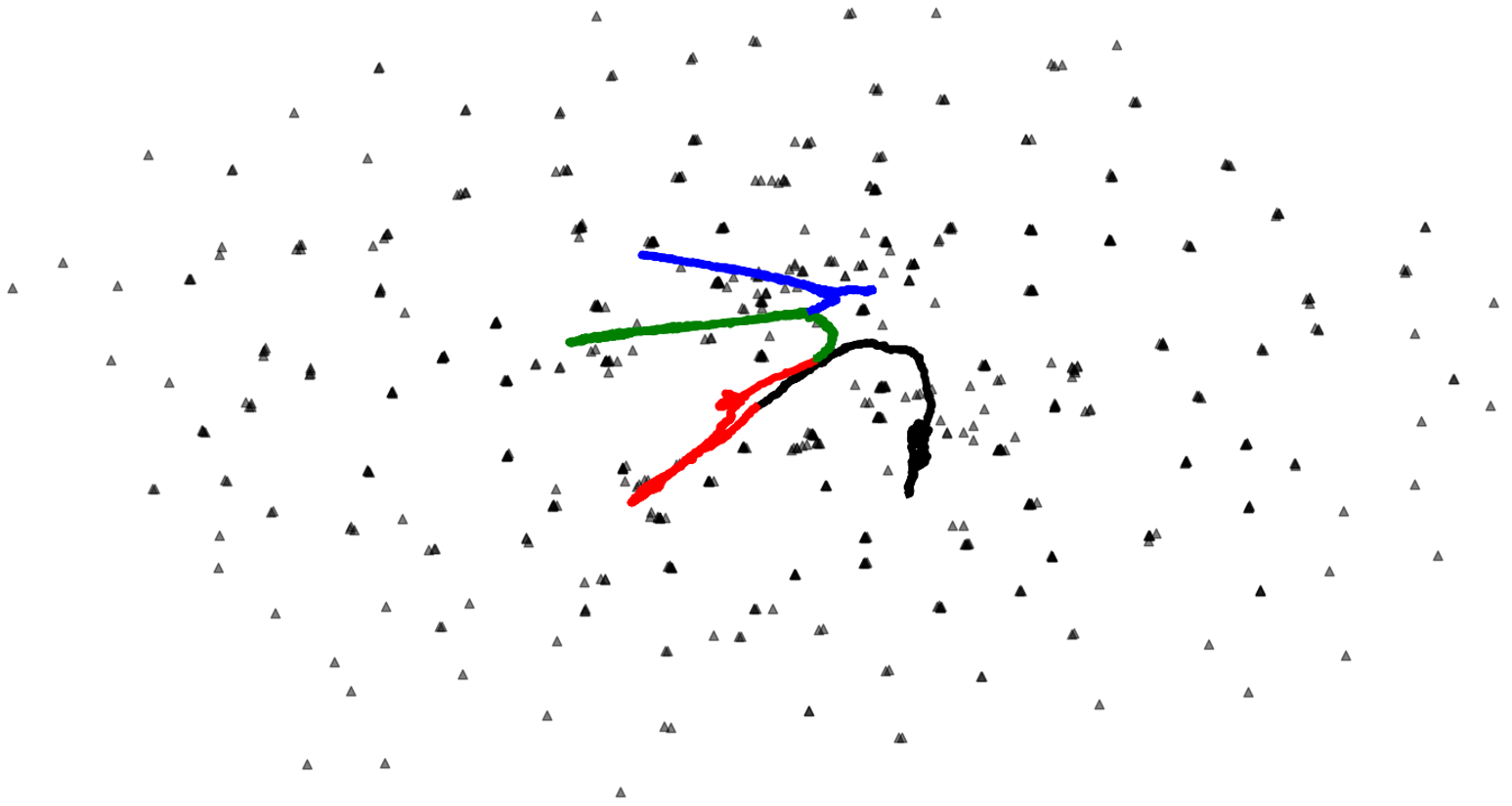}
        \caption{Human Associations}
    \end{subfigure}
    ~
    \begin{subfigure}{0.23\textwidth}
        \includegraphics[width=\columnwidth]{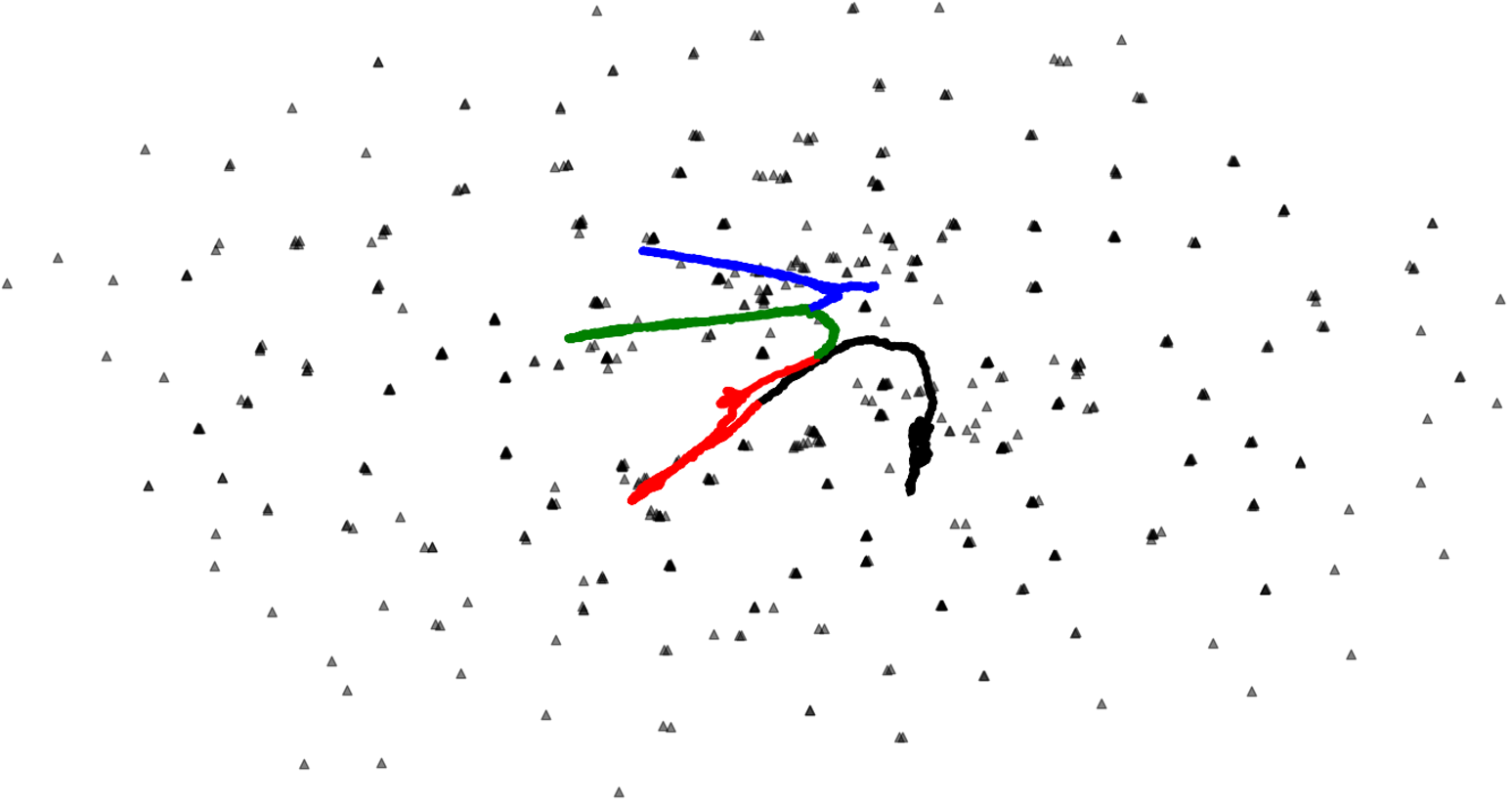}
        \caption{Ours}
    \end{subfigure}
    ~
    \begin{subfigure}{0.23\textwidth}
        \includegraphics[width=\columnwidth]{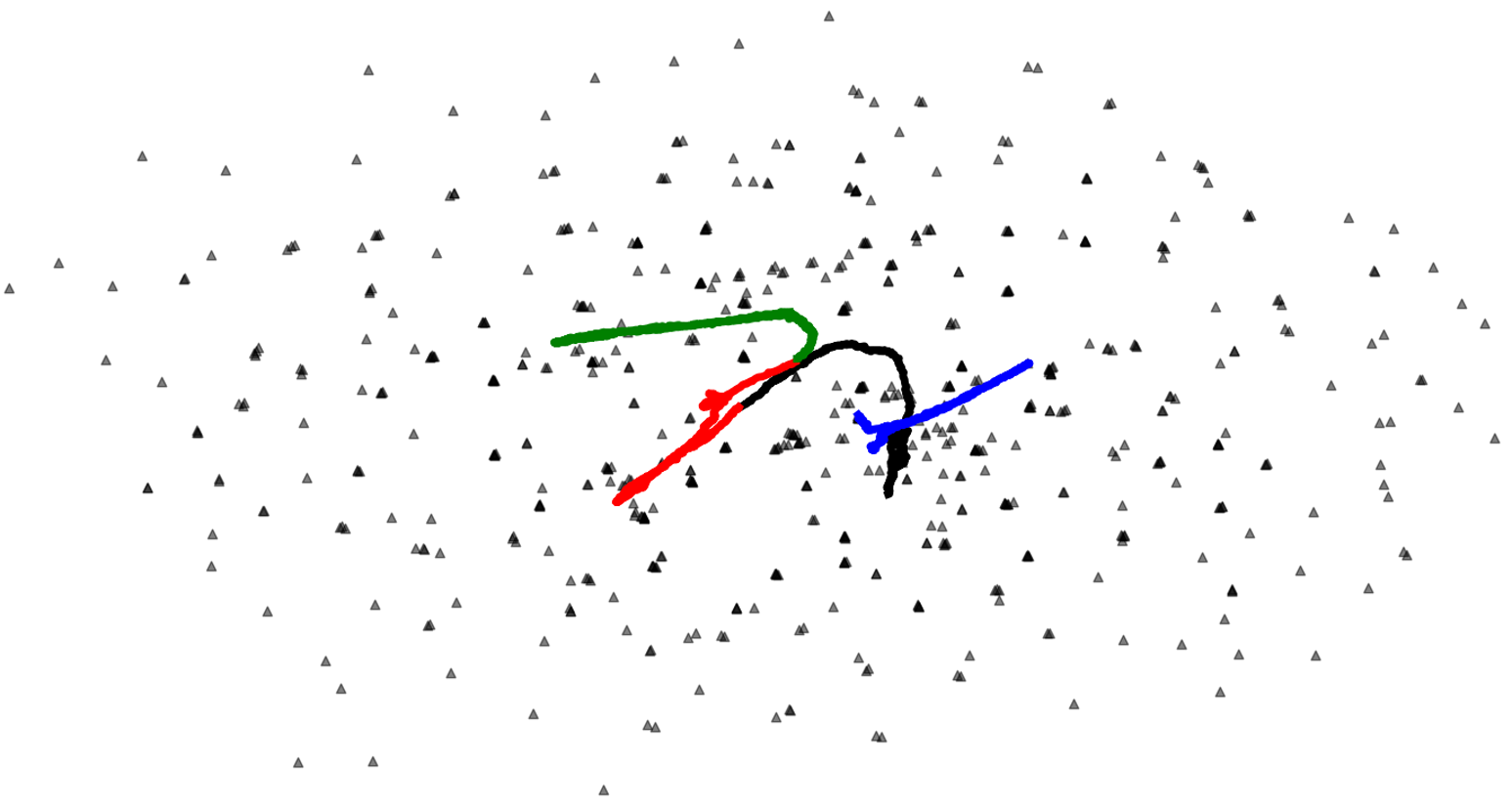}
        \caption{GLAROT~\cite{kallasi2016efficient}}
    \end{subfigure}
    ~
    \begin{subfigure}{0.23\textwidth}
        \includegraphics[width=\columnwidth]{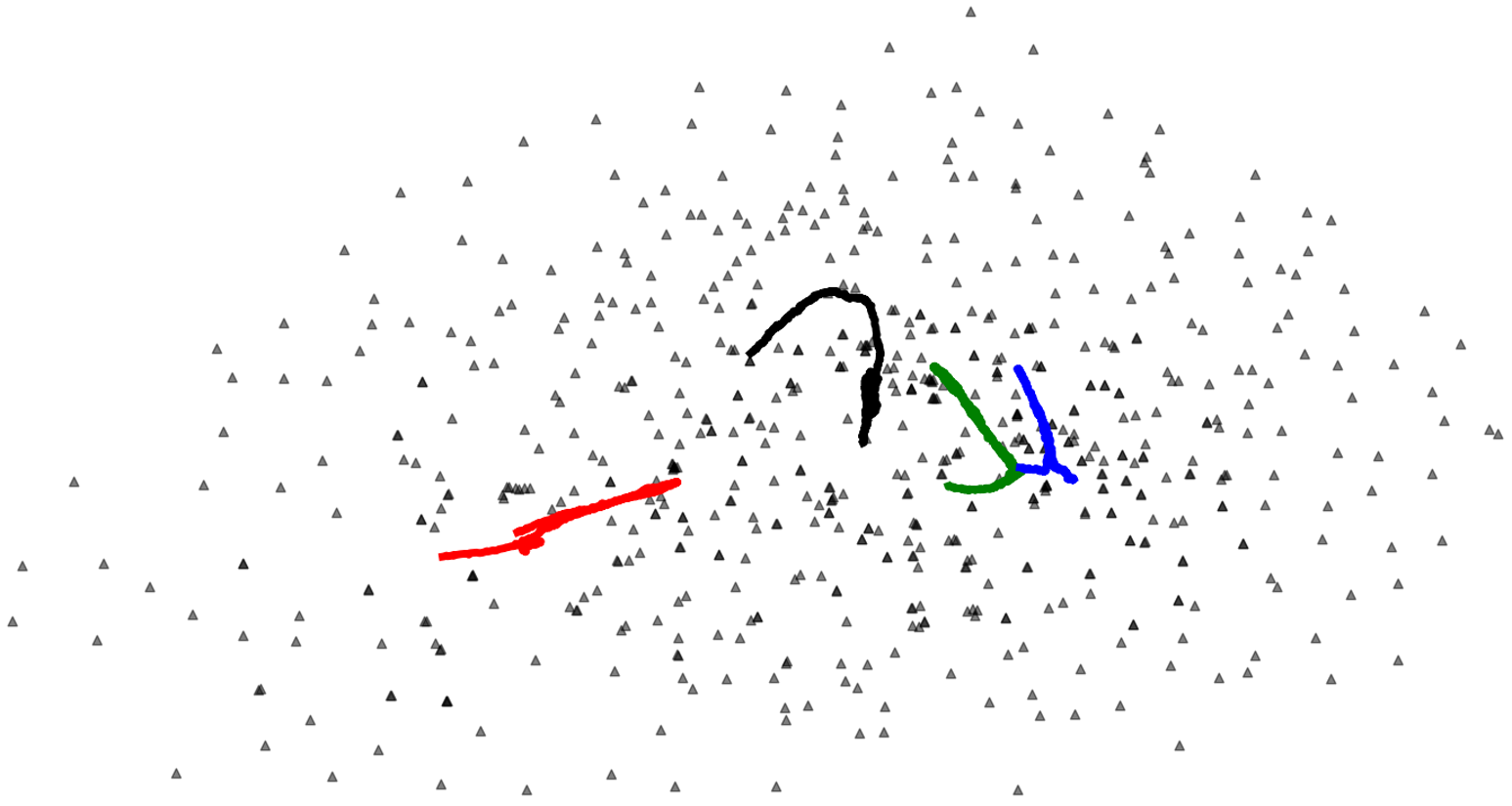}
        \caption{Li et al.~\cite{li2020localization}}
    \end{subfigure}
    \caption{Accumulated sub-maps for flight one (top row) and flight two (bottom row). Colors represent the robot pose in different sub-maps, and black triangles represent the landmarks. We iteratively merge pairs of sub-maps until the entire trajectory is accumulated into a single map. Our method closely approximates to the results obtained with human associations.}
    \label{fig:submap_merging_trajectories}
    \vspace{-5mm}
\end{figure*}
\begin{table}
\centering
\caption{Landmark alignment error. We compare the distance between manually associated landmarks after the alignment with correspondences provided by each method.}
\resizebox{\columnwidth}{!}{%
  \begin{tabular}{|l|ccc|ccc|}
    \hline
    \multirow{2}{*}{\backslashbox{Method}{Experiment}} &
      \multicolumn{3}{c|}{Flight 1} &
      \multicolumn{3}{c|}{Flight 2} \\
    & Mean & Min. & Max. & Mean & Min. & Max. \\
    \hline
    Human Associations & $0.17$ & $0.00$ & $0.54$ & $0.21$ & $0.01$ & $0.67$ \\\hline
    Ours & $\mathbf{0.19}$ & $\mathbf{0.00}$ & $\mathbf{0.61}$ & $\mathbf{0.23}$ & $\mathbf{0.01}$ & $\mathbf{0.83}$ \\
    GLAROT~\cite{kallasi2016efficient} & $0.20$ & $0.01$ & $0.90$ & $30.15$ & $0.03$ & $89.22$ \\
    Li et al.~\cite{li2020localization} & - & - & - & $42.38$ & $4.79$ & $125.71$ \\\hline
\end{tabular}%
}
\label{tab:map_merging_results}
\end{table}

In this experiment, we fly the robot twice across the same plot of a commercial pine tree forest. For each flight, we split the raw sensor readings into sub-maps for every minute of data. We run SLOAM~\cite{chen2020sloam} for pose estimation and landmark detection on each sub-map individually. The sub-maps have partial overlap, position noise, and detection failure cases, making this task challenging for methods that are not robust to these factors.

We match sub-maps following a chronological order and iteratively recompute a descriptor for the accumulated map to merge with the next sub-map until all are combined into a single map. For GLAROT~\cite{kallasi2016efficient}, since we know that subsequent sub-maps have overlap, we skip the global \textcolor{black}{descriptor} check and directly match landmarks.

The associations could be used in a more complex data association pipeline such as CLEAR~\cite{fathian2019clear} that checks for cycle consistency across different sub-maps to refine the matches resulting in better map quality. In this experiment, we use a simpler approach with DBSCAN~\cite{ester1996density} to cluster trees closer than $0.5m$ after the alignment into a single landmark. This method is preferable instead of removing duplicates based on the correspondences since some landmark matches may not have been detected.

\textcolor{black}{
The forest canopy causes GPS to have errors that can range up to tens of meters, which prevents it from being used as ground truth.
Although the odometry provided by SLOAM~\cite{chen2020sloam} may contain drift, from our results in previous work, we observed errors smaller than $1$ meter for trajectories of similar length in a similar forest. Besides the possible drift, since each sub-map is an independent SLOAM run, the trees will not be perfectly aligned. Instead, we use it as an initial guess to align subsequent sub-maps, manually annotate tree correspondences between pairs of sub-maps, and use RANSAC to compute a transformation.
}

\textcolor{black}{
For all methods, we compute a translation error based on the Euclidean distance, and a rotation error based on the absolute difference for each transformation between subsequent sub-maps when compared with the transformations derived from the human associations. For our method, the translation error on flight 1 is $0.43$ meters, and the rotation error is $0.32$ degrees. For GLAROT~\cite{kallasi2016efficient}, the translation and rotation errors are $0.33$ meters and $1.54$ degrees, respectively, and Li et al.~\cite{li2020localization} fails to find correspondences for one of the sub-maps. On flight two, the translation and rotation errors are $0.26$ meters and $0.33$ degrees for our method, while the others have translation errors greater than $15$ meters and $200$ degrees. In Fig.~\ref{fig:submap_merging_trajectories}, we show that on flight 1, our method and GLAROT~\cite{kallasi2016efficient} closely approximates the result with human associations, and Li et al.~\cite{li2020localization} also approximates it except for the sub-map that it fails to align. On flight 2, our method is similar to the reference, while the benchmarks are significantly off.
}

\textcolor{black}{
With the transformations computed with manual annotations, we select tree associations considered inliers by RANSAC to evaluate the quality of landmark alignments, totaling $142$ and $162$ trees for flights 1 and 2 respectively. The distance between these landmarks after alignment is computed on pairs of subsequent sub-maps. In Table~\ref{tab:map_merging_results}, we show the mean, minimum and maximum distances over all pairs compared to the human associations. Similar to our previous results, our method closely approximates the reference on both flights, while GLAROT~\cite{kallasi2016efficient} achieves good results in flight one but fails in flight two, and Li et al.~\cite{li2020localization} fails in both flights.
}

\subsection{Efficiency}
For deployment on a robot, an algorithm has to be accurate and have a reasonable execution time. For all three methods, ~\textcolor{black}{we run offline tests} and report in \textcolor{black}{Table}~\ref{tab:times} the worst-case big O and the average speed required to estimate landmark correspondences between two observations. ~\textcolor{black}{Times are reported using an Intel® i7-7500U CPU, while the UAV has a more powerful Intel® i7-8559U.
}

Even though our method has quadratic big O for matching in the worst case, it scales with the number of polygons of $\mathcal{H}^{2}(t)$, which is smaller than the number of landmarks in a given observation. Moreover, we can sample polygons to have an upper bound on matching time if necessary.

For Li et al.~\cite{li2020localization}, the complexity for computing the descriptors is bounded by the computation of the Delaunay triangulation. However, as stated in Sec.~\ref{ssec:polys}, the number of triangles can be larger than the number of points, which can be prohibitive during matching.

It is worth noting that our method and Li et al.~\cite{li2020localization} are implemented in pure Python. For GLAROT~\cite{kallasi2016efficient}, we ported the original author's C++ code to Python. However, during our experiments, we run the \textcolor{black}{descriptor} computation in pure Python and matching in C++ with Python bindings due to the high execution time of the Python version.

\begin{table}[!ht]
\caption{Worst case big O and time benchmarks for feature computation and matching for a pair of observations $i$ and $j$. $n_i$ is the number of landmarks in $i$, and $g_i$ is the number of polygons in $i$. Reported times are the median time per observation in milliseconds.}
\begin{tabular}{l|p{1.5cm}p{1.2cm}|p{1cm}p{1.2cm}}
\multirow{2}{*}{} &
\multicolumn{2}{c}{Descriptor computation} &
\multicolumn{2}{c}{Matching} \\
    & Big O & Time & Big O & Time \\
Ours (Py)  & $(n \, log \, n) + l$ & $16$ & $g_i * g_j$ & $1.24$ \\
GLAROT~\cite{kallasi2016efficient} (Py) & $n^2$ & $61$ & $n_i*n_j$ & $2.55$\\
GLAROT~\cite{kallasi2016efficient} (C++) & $n^2$ & $4$ & $n_i*n_j$ & $0.007$\\
Li et al.~\cite{li2020localization} (Py)  & $(n \, log \, n)$ & $15$ & $g_i * g_j$ & $47$
\end{tabular}

\label{tab:times}
\vspace{-4mm}
\end{table}

%% file: tex/conc.tex
Place recognition and data association are challenging problems. Especially in a forest where trees that look identical are the only available landmarks.

We presented a method that defines sets of polygons based on tessellations computed on the position of landmarks and a framework that uses these polygons to identify previously seen locations and compute landmark associations.

Our experiments show that the proposed method is more reliable and robust compared to the benchmarks. The construction of polygons from triangles narrows the search space of possible landmarks correspondences. When combined with consolidated shape retrieval techniques for matching, these polygons yield a more robust framework than relying purely on geometric properties such as polygon area and perimeter.

Using data captured from a UAV in a real forest, we show the advantages of having reliable features that describe only parts of the observation while merging sub-maps with partial overlap and noise. We manually annotated landmark associations in subsequent sub-maps and compared all methods against the alignments obtained with these associations. 
Our method outperformed both benchmarks in most metrics and closely approximated the reference when looking at the transformations directly. We also use manual landmark associations to compare the quality of the resulting map, where again, our method was the closest to the reference map.

One of the drawbacks of using local structures without a global ~\textcolor{black}{descriptor}, such as GLARE, is that the number of elements to match grows significantly faster as we accumulate observations for tasks such as loop closure. Our method reduces the search space compared with local methods yet still grows in computational cost with the number of landmarks. We presented a solution based on sampling polygons but show that it decreases the performance of the method.

\textcolor{black}{
Since our framework's input is a set of landmark positions projected on 2-D, it can also be used for associating observations from over and under canopy data as long as these detections are partially consistent. This is valuable since it is easier to estimate the trees' diameter from under the canopy, while over canopy height estimation is a well-studied problem~\cite{pang2008forest}.} Other directions for future work include \textcolor{black}{testing in denser and more cluttered forests}, better polygon sampling techniques, an extension of the descriptor computation and matching to 3D shapes, and ways to incorporate other properties, such as dimensions of the objects to improve accuracy.